\documentclass[letterpaper, 10 pt, journal, twoside]{IEEEtran}
\ifCLASSINFOpdf
\else
\fi
\usepackage{graphicx}
\usepackage{graphics}
\usepackage{times}
\usepackage{amsmath}
\usepackage{amssymb}
\usepackage{float}
\usepackage{url}
\usepackage{subcaption}
\usepackage{caption}
\usepackage{multirow}
\usepackage{gensymb}
\usepackage{epstopdf}
\usepackage{wrapfig}
\usepackage{setspace} 
\usepackage{algorithm}
\usepackage{amsmath,amssymb}
\usepackage{algorithm}
\usepackage{algpseudocode} 
\algrenewcommand\algorithmicrequire{\textbf{Input:}}
\algrenewcommand\algorithmicensure{\textbf{Output:}}

\usepackage{color}
\usepackage{cite}
\usepackage{booktabs}
\usepackage{stackengine}
\usepackage[font={small}]{caption}
\usepackage[multiple]{footmisc}

\usepackage{tikz}
\usepackage{textcomp}
\usepackage{lipsum}
\usepackage{mathtools}
\usepackage[multiple]{footmisc}
\usepackage{cuted}
\usepackage{booktabs}
\usepackage{ulem}   
\usepackage{xspace}
\usepackage{hyperref}
\usepackage{cleveref}
\crefname{section}{Sec.}{Secs.}
\crefname{algorithm}{Alg.}{Algs.}
\crefname{figure}{Fig.}{Figs.}
\Crefname{section}{Section}{Sections}
\usepackage{comment}

\hypersetup{
    colorlinks=true,
    linkcolor=blue,
    filecolor=magenta,      
    urlcolor=cyan,
}

\DeclareMathOperator*{\argmin}{argmin}
\newcommand{\myParagraph}[1]{%
  \textbf{#1.}\xspace%
}

\def\name{LatentAM\xspace}
\title{\LARGE \bf LatentAM: Real-Time, Large-Scale Latent Gaussian Attention Mapping via Online Dictionary Learning
}

\usepackage[table]{xcolor}
\definecolor{bestgreen}{RGB}{198,239,206}
\definecolor{secondyellow}{RGB}{255,235,156}

\newcommand{\best}[1]{\cellcolor{bestgreen}{#1}}
\newcommand{\second}[1]{\cellcolor{secondyellow}{#1}}

\newif\ifrevision
\revisionfalse   

\newcommand{\rev}[1]{%
  \ifrevision
    \textcolor{blue}{#1}%
  \else
    #1%
  \fi
}

\definecolor{revblue}{RGB}{0,70,180}

\begin{document}
%

\def\name{LatentAM\xspace}
\title{LatentAM: Real-Time, Large-Scale Latent Gaussian Attention Mapping via Online Dictionary Learning
}

%
%
%

\author{Junwoon Lee and Yulun Tian}

%
%

\markboth{IEEE Robotics and Automation Letters. Preprint Version. Accepted July, 2026}
{Lee \MakeLowercase{\textit{et al.}}: LatentAM: Real-Time, Large-Scale Latent Gaussian Attention Mapping via Online Dictionary Learning} 

%




\twocolumn[{%
\renewcommand\twocolumn[1][]{#1}%

\maketitle

\begin{center}
  \centering
  \vspace{-17pt}
  \captionsetup{type=figure}
  \includegraphics[width=0.99\linewidth]{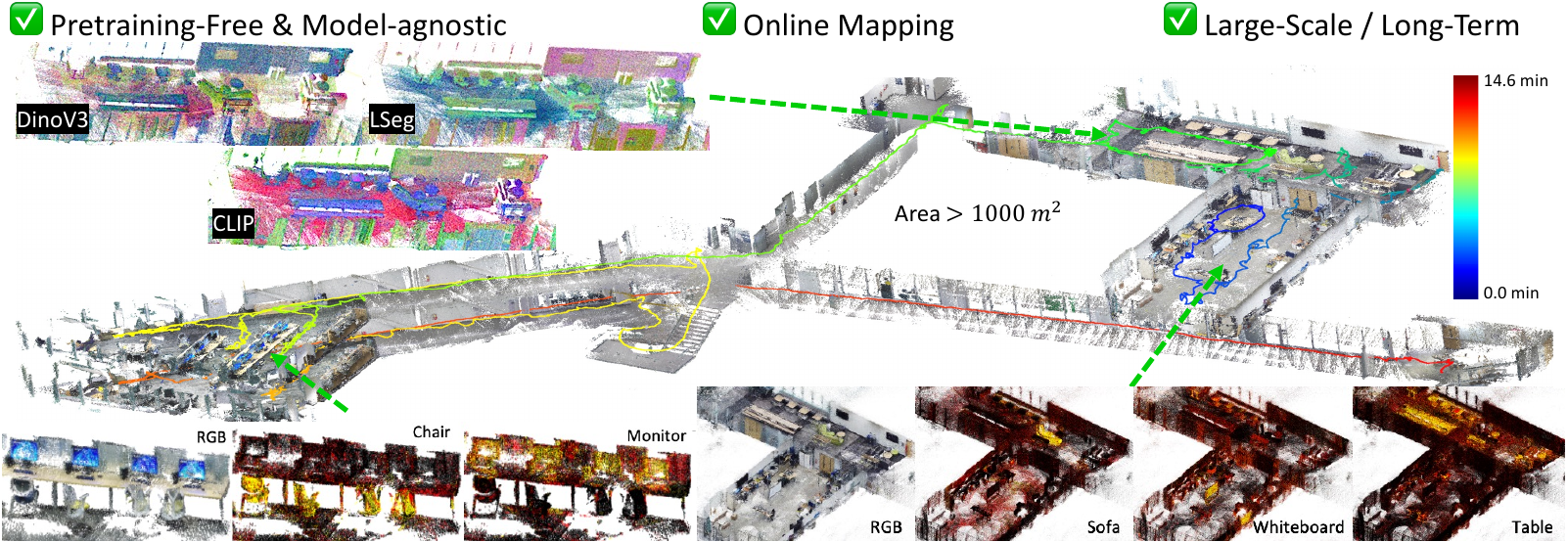}
  \setlength{\abovecaptionskip}{6pt}
  \captionof{figure}{Large-scale online latent mapping with \textbf{\name} on a custom multi-floor dataset spanning approximately 530\,m of trajectory over 14.6 minutes.
  The system incrementally constructs a latent 3D Gaussian map from streaming RGB-D observations, while supporting plug-and-play integration with different visual–language models including CLIP~\cite{radford2021learning}, DINOv3~\cite{simeoni2025dinov3}, and LSeg~\cite{li2022language}.
  The resulting latent map directly supports downstream open-vocabulary perception tasks such as language-driven querying in 3D. }
  \label{fig:mot}
  \vspace{0pt}
\end{center}%
}]

\makeatletter
\renewcommand*{\@makefnmark}{}
\makeatother

\footnotetext{
Manuscript received: February 6, 2026; Revised: May 4, 2026; Accepted: July 23, 2026.
}

\footnotetext{
This paper was recommended for publication by Editor Javier Civera upon evaluation of the Associate Editor and Reviewers' comments.
}

\footnotetext{
Junwoon Lee and Yulun Tian are with the Robotics Department, University of Michigan, Ann Arbor, MI, United States.
{\tt\footnotesize \{junwoon, yulunt\}@umich.edu}
}

\footnotetext{
Digital Object Identifier (DOI): see top of this page.
}

\begin{abstract}
We present \name, an online 3D Gaussian Splatting (3DGS) mapping framework that builds scalable latent feature maps from streaming RGB-D observations for open-vocabulary robotic perception. Instead of distilling high-dimensional vision-language Model (VLM) embeddings using model-specific decoders, \name proposes an online dictionary learning approach that is both model-agnostic and pretraining-free, enabling plug-and-play integration with different VLMs at test time.
Specifically, our approach associates each Gaussian primitive with a compact query vector that can be converted into approximate VLM embeddings using an attention mechanism with a learnable dictionary.
The dictionary is initialized efficiently from streaming observations and optimized online to adapt to evolving scene semantics under trust-region regularization.
To scale to long trajectories and large environments, we further propose an efficient map management strategy based on voxel hashing, where optimization is restricted to an active local map on the GPU, while the global map is stored and indexed on the CPU to maintain bounded GPU memory usage.
Experiments on public benchmarks and a large-scale custom dataset demonstrate that \name attains significantly better feature reconstruction fidelity compared to state-of-the-art methods, while achieving near-real-time speed (12-35~FPS) on the evaluated datasets. Our project page, including the code, is available at: \url{https://junwoonlee.github.io/projects/LatentAM}
\end{abstract}


%
\IEEEpeerreviewmaketitle


\section{INTRODUCTION}

\IEEEPARstart{S}{imultaneous}
Localization and Mapping (SLAM) is a cornerstone ability that provides situational awareness for mobile robots in unknown environments.
Recent advances have significantly improved robustness, scalability, and integration with downstream tasks. 
However, emerging robotic applications require richer spatial grounding than what is supported by existing geometry-centric or closed-set semantic SLAM approaches, demanding fine-grained and general-purpose scene understanding. 
To this end, recent vision-language models (VLMs) \cite{radford2021learning,simeoni2025dinov3,li2022language, kirillov2023segment, touvron2023llama} 
offer powerful visual priors learned from internet-scale data.
A central challenge, however, is to effectively deploy such representations within onboard SLAM, where perception must be performed incrementally from streaming observations under occlusions, limited viewpoints, and strict runtime and memory constraints.

To address the challenges posed by occlusions and viewpoint variability, recent work has explored fusing 2D VLM embeddings into consistent 3D representations based on point clouds~\cite{conceptfusion},  neural radiance field \cite{kerr2023lerf, weijler2025openhype}, volumetric maps \cite{wilson2025latent}, and 3D Gaussian Splatting (3DGS) \cite{qiu2024feature, zhou2024feature,zou20253d, wu2024opengaussian, yang2025opengs, yang2025opengsfusion, titkov2025leg, lee2025lego, yoo2025openmonogs, tian2025ccl, qu2024goi}.
Recently, 3DGS \cite{kerbl20233d} has emerged as an attractive representation due to its superior real-time rendering performance.
While the aforementioned approaches typically focus on offline reconstruction,
a growing body of work \cite{katragadda2025online,titkov2025leg,lee2025lego,deng2025omnimap,yoo2025openmonogs} integrates VLM embeddings with
incremental 3DGS mapping to enable 3D visual–language mapping from streaming observations.
Despite this progress, existing approaches exhibit two critical limitations. 
First, they remain computationally expensive, often operating far below sensor rate and limited to small-scale (e.g., single room) environments.
Second, many methods rely on feature distillation by pretraining model-specific encoders or decoders. Such a design tightly couples the mapping system to specific choices of VLM, making it costly to incorporate new models or adapt to evolving feature distributions.

\textbf{Contributions.}
We present \name, a novel framework for online visual–language mapping based on feature-augmented 3D Gaussian Splatting from streaming RGB-D observations. 
Rather than relying on feature distillation that requires model-specific encoders or decoders, our core insight is to formulate online feature mapping as a \emph{dictionary learning} \cite{mairal2009online} problem, in which a compact dictionary of VLM embeddings are incrementally constructed and updated as the robot navigates and maps the environment. 
The online reconstructed dictionary enables our approach to continuously adapt to new environments, viewpoints, and semantic content, while a trust-region-style regularization prevents overfitting and catastrophic forgetting during long-term operation. 
As a direct consequence, our approach is both \emph{model-agnostic} and \emph{pretraining-free};
in our experiments, we demonstrate plug-and-play integration with CLIP~\cite{radford2021learning}, DINOv3~\cite{simeoni2025dinov3}, and LSeg~\cite{li2022language} within the same framework.
Furthermore, by integrating the proposed dictionary-based feature learning into a scalable local–global 3DGS mapping system with voxel hashing, we achieve substantially improved mapping speed and memory efficiency, enabling near-real-time operation (12--35 FPS depending on parameter settings) and scaling to large-scale environments ($>$ 530~m) as shown in \Cref{fig:mot}. 
The proposed method improves feature reconstruction accuracy by 39.2\% and achieving 28.8$\times$ speedups compared to the state-of-the-art method~\cite{zou20253d}.

\section{Related Work}
\subsection{SLAM with 3D Gaussian Splatting (3DGS)} 
3D Gaussian Splatting~\cite{kerbl20233d} has recently emerged as a powerful representation for high-fidelity real-time 3D reconstruction and rendering, motivating 3DGS-based SLAM systems~\cite{matsuki2024gaussian, keetha2024splatam, ha2024rgbd, yang2025opengs, yang2025opengsfusion}; see also~\cite{tosi2024nerfs}.

GS-ICP~\cite{ha2024rgbd} introduces a 3DGS  SLAM framework that achieves real-time performance by using fast Gaussian primitive registration via geometric alignment. 
OpenGS-SLAM~\cite{yang2025opengs} incorporates semantic labels into a GS-ICP backbone to ensure semantic consistency, 
while OpenGS-Fusion~\cite{yang2025opengsfusion} extends this framework to a hybrid representation based on 3DGS and truncated signed distance function.

\subsection{Offline Feature Mapping} 
Several works bridge 2D VLMs and 3D spatial representations by fusing foundation-model embeddings into neural 3D representations. 
LERF~\cite{kerr2023lerf} pioneered language-embedded neural representations by volume-rendering CLIP features along NeRF rays with multi-scale CLIP supervision. 
Several works extend this paradigm to 3DGS by distilling high-dimensional foundation-model embeddings into compact 3DGS feature representations~\cite{qiu2024feature, zhou2024feature, zuo2025fmgs, qin2024langsplat, shi2024language}. 
Feature-splatting~\cite{qiu2024feature} supervises 3D Gaussians with photometric and feature reconstruction losses, enabling downstream tasks such as segmentation, editing, and retrieval. 
M3~\cite{zou20253d} proposes a different paradigm that avoid direct distillation of high-dimensional embeddings into Gaussians. Instead, it stores observed embeddings in a global dictionary as principal scene components and learns low-dimensional queries for each Gaussian via attention. 
OpenGaussian~\cite{wu2024opengaussian} associates 3D points with 2D instance-level features in a spatial codebook, enabling position-dependent lustering of semantic features. 
Recently, GOI~\cite{qu2024goi} and CCL-LGS~\cite{tian2025ccl} propose learnable codebooks trained with clustering-based objectives, yielding compact atoms that improve reconstruction quality under feature distribution shifts. 
However, these methods rely on offline learning of the codebook, require substantial memory for clustering-based updates, and lack incremental dictionary maintenance for streaming settings.

\subsection{Incremental and Online Feature Mapping} 
As an online feature mapping method, Online Language Splatting~\cite{katragadda2025online} enables online open-vocabulary mapping within an 3DGS-SLAM pipeline by distilling CLIP features through a lightweight super-resolution decoder. 
Similarly, LEGO-SLAM~\cite{lee2025lego} learns a scene-adaptive decoder to distill high-dimensional features into low-dimensional Gaussian embeddings, and further employs a latent feature codebook that remains fixed after construction. 
LEG-SLAM~\cite{titkov2025leg} adopts a different strategy by embedding VLM features using a pretrained PCA projection. 
Similar to~\cite{katragadda2025online}, 
FeatureSLAM~\cite{thirgood2026featureslam} introduces a pretrained autoencoder with lightweight learnable adaptation layers to efficiently handle streaming data, while improving tracking stability via a feature-based pose tracker. 
However, distillation-based methods are highly sensitive to model capacity, where insufficient capacity leads to underfitting and excessive capacity increases memory usage and training time. 

To address the capacity-accuracy trade-off of decoder-based distillation, 
recent works have explored codebook-based schemes for efficiently reconstructing high-dimensional embeddings in incremental mapping. 
Following OpenVox~\cite{deng2025openvox}, OmniMap~\cite{deng2025omnimap} maintains a global embedding codebook indexed by semantic instance IDs, where each atom stores an instance-level embedding updated over time. 
OpenMonoGS-SLAM~\cite{yoo2025openmonogs} similarly maintains a compact codebook of representative embeddings, updates it online using pairwise cosine similarity among atoms, and reconstruct high-dimensional embeddings via memory attention~\cite{zou20253d}. 
OpenFusion++~\cite{jin2025openfusion++} proposes a TSDF-based open-vocabulary mapping framework with instance-level semantics, where embeddings are incrementally updated through a semantic cache that functions as a codebook. 
Nevertheless, by fixing the dictionary atoms, these methods cannot adapt their latent basis to the non-stationary feature distribution encountered over long trajectories, causing stale atoms as new semantics appear and ultimately limiting large-scale mapping scalability.

\section{Proposed Method}

\subsection{Preliminary: 3DGS Representation}
\label{sec:prelim_3dgs}

We represent a scene as a collection of 3D Gaussian primitives
$\mathcal{G}=\{G_i\}_{i=1}^N$,
where each primitive encodes geometric and appearance attributes: 
\begin{equation}
G_i =
\left\{
\mathbf{p}_i,\;
\mathbf{q}_i,\;
\mathbf{c}_i,\;
\mathbf{s}_i,\;
\alpha_i
\right\}.
\label{eq:gs_repr_basic}
\end{equation}
Here, $\mathbf{p}_i\in\mathbb{R}^3$ denotes the position,
$\mathbf{q}_i\in\mathbb{R}^4$ the orientation,
$\mathbf{c}_i\in\mathbb{R}^3$ the color,
$\mathbf{s}_i\in\mathbb{R}^3$ the scale, and
$\alpha_i\in[0,1]$ the opacity. 
We use view-independent color and omit spherical harmonics to enable efficient online optimization like~\cite{matsuki2024gaussian}.

The renderer $\mathcal{R}(\cdot)$ maps a set of Gaussian primitives $\mathcal{G}$ to an
image using depth-ordered alpha blending~\cite{kerbl20233d}.
Let $\{G_i\}_{i=1}^N$ denote the Gaussians intersecting a given pixel ray $p$, sorted by increasing depth.
The rendered color is,
\begin{equation}
\mathbf{C}_p
=
\sum_{i=1}^{N}
\alpha_i\,\mathbf{c}_i
\prod_{j=1}^{i-1}
\left(1-\alpha_j\right),
\label{eq:alpha_blending}
\end{equation}
A per-pixel depth value can be rendered using the same alpha-blending operation. 
Given the camera pose $\mathbf{T}\in\mathrm{SE}(3)$ and known camera intrinsics, we denote the rendering of color and depth by
$\widehat{\mathbf{I}} = \mathcal{R}_c(\mathcal{G}, \mathbf{T})$ and
$\widehat{\mathbf{Z}} = \mathcal{R}_d(\mathcal{G}, \mathbf{T})$,
where $\widehat{\mathbf{I}}$ and $\widehat{\mathbf{Z}}$ are the rendered 2D color image and depth.

\begin{figure}[t]
	\centering
	\includegraphics[width=.99\columnwidth]{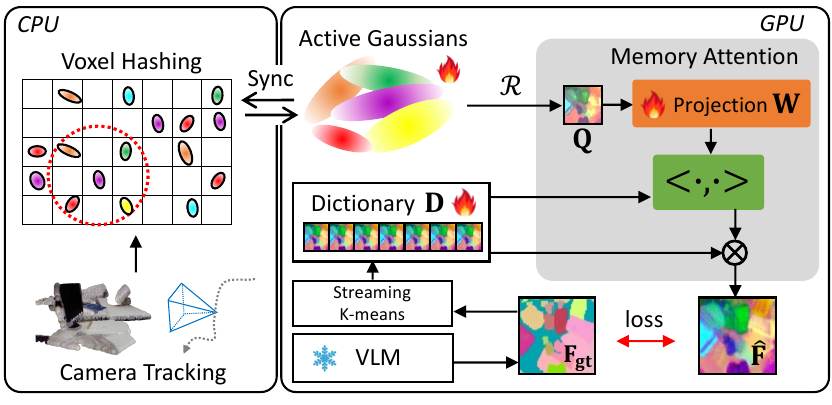}
	\caption{Overview of our system, including camera tracking, voxel-hashing-based global map, and online dictionary learning.}
	\label{fig:systemoverview}
	\vspace{-5mm}
\end{figure}

\subsection{Feature Splatting via Dictionary Learning}
\label{sec:feature_splatting}

In this subsection, we extend standard 3DGS to scalable \emph{latent feature mapping} as shown in Fig.~\ref{fig:systemoverview}.
Instead of storing high-dimensional features at each Gaussian, 
each primitive stores a low-dimensional query vector, 
from which full embeddings are reconstructed via dictionary learning.

\myParagraph{Feature-augmented Gaussian Representation}
In addition to the geometric and photometric attributes introduced in
Sec.~\ref{sec:prelim_3dgs}, each Gaussian primitive stores a latent query feature
$\mathbf{f}_i\in\mathbb{R}^{d_s}$, i.e., 
\begin{equation}
G_i =
\left\{
\mathbf{p}_i,\;
\mathbf{q}_i,\;
\mathbf{c}_i,\;
\mathbf{s}_i,\;
\alpha_i,\;
\mathbf{f}_i
\right\}.
\end{equation}
Here, $d_s \ll d_f$ is the dimension of the query vector (set to $32$ by default in our implementation), and $d_f$ is the dimension of the target VLM feature, 
e.g., $d_f = 768$ for CLIP \cite{radford2021learning}. 
Given camera pose $\mathbf{T}$, 
we render a per-view query feature map by applying the alpha blending operator~\eqref{eq:alpha_blending} 
to query features $\mathbf{f}_i$ instead of colors $\mathbf{c}_i$. 
In the following, we denote this operation by 
\begin{equation}
\mathbf{Q} = \mathcal{R}_f(\mathcal{G}, \mathbf{T}) \in \mathbb{R}^{n \times d_s}.
\label{eq:render_query}
\end{equation} 
For SAM-based extraction, $n$ denotes the number of valid SAM~\cite{kirillov2023segment, mobile_sam} masks, and each feature vector is obtained by mask-pooling normalized CLIP/DINO embeddings within the corresponding region, following~\cite{qiu2024feature}.

\myParagraph{Embedding Reconstruction with Dictionary Learning} 
Existing latent mapping techniques typically rely on 3D feature distillation, 
which compresses high-dimensional foundation model embeddings into low-dimensional Gaussian attributes via learned decoders~\cite{zhou2024feature, qiu2024feature}.
While effective for reducing memory and computation, such distillation introduces an information bottleneck, causing loss of semantic fidelity and misalignment between the decoded embeddings and the original targets, as confirmed in (\cref{sec:experiment:ablation}). 

In contrast, the proposed approach builds on a dictionary-based strategy that is first proposed in M3~\cite{zou20253d}.
Specifically, we introduce and learn (i) a projection matrix $\mathbf{W}\in\mathbb{R}^{d_s\times d_f}$, and (ii) a memory bank (dictionary) $\mathbf{D}\in\mathbb{R}^{K\times d_f}$ that stores $K$ representative embeddings in the target VLM space. 
Notably, the additional memory cost of $\mathbf{W}$ and $\mathbf{D}$ remains very small compared to the memory footprint of the Gaussian primitives, adding about 3-6\% in typical single-room scenes. 
Using $\mathbf{W}$ and $\mathbf{D}$, we convert the queries $\mathbf{Q}$ in \eqref{eq:render_query} into approximate embeddings via, 
\begin{equation}
\widehat{\mathbf{F}}
=
\mathrm{Softmax}\!\left( \mathbf{Q} \mathbf{W}\mathbf{D}^\top\right)\mathbf{D}.
\label{eq:dict_recon}
\end{equation}
Following M3~\cite{zou20253d}, we compute the softmax term in \eqref{eq:dict_recon} using the cosine similarity between $\mathbf{Q}\mathbf{W}$ and $\mathbf{D}$ after $\ell_2$ normalization and with temperature scaling. 
We note that \eqref{eq:dict_recon} can be viewed as augmenting the standard feature decoder with an attention-based reconstruction mechanism.
In particular, the matrix $\mathbf{W}$ plays the role of a minimal linear decoder that maps $\mathbf{Q}$ to an initial reconstruction $\widetilde{\mathbf{F}} = \mathbf{Q}\mathbf{W} \in \mathbb{R}^{n\times d_f}$. 
Then, the softmax operation in \eqref{eq:dict_recon} computes pairwise similarities between $\widetilde{\mathbf{F}}$ and $\mathbf{D}$, yielding attention weights that reflect the relevance of each dictionary atom. The final reconstructed embedding $\widehat{\mathbf{F}}$ is obtained as a weighted combination of dictionary atoms.
This approach enables more expressive embeddings and avoids the information bottleneck inherent to direct distillation.

Nevertheless, the original approach in \cite{zou20253d}
assumes that the full set of latent embeddings is available \emph{a priori}, and its memory bank is constructed in an offline manner. 
Thus, the approach does not naturally extend to streaming settings, where observations arrive sequentially and feature distributions evolve over time. 
To address this, we take inspiration from the established research field of dictionary learning~\cite{mairal2009online} and proposes a {regularized optimization} formulation that explicitly handles non-stationary feature distribution while preventing overfitting. 
Specifically, we jointly optimize the Gaussian primitives $\mathcal{G}$, the projection matrix $\mathbf{W}$, and the dictionary $\mathbf{D}$ via,
\begin{equation}
\begin{aligned}
\min_{\mathcal{G},\,\mathbf{W},\,\mathbf{D}}\;\;
\mathcal{L}_{\mathrm{f}}
=&\;
\lambda_{\mathrm{cos}}
\bigl(1-\mathrm{cos}(\widehat{\mathbf{F}},\mathbf{F})\bigr)
+
\lambda_{2}\|\widehat{\mathbf{F}}-\mathbf{F}\|_2^2 \\
&+
\lambda_D \sum_{j=1}^{K}
\mathrm{ReLU}\!\left(
\left\|
\mathbf{D}^{(0)}_j - \mathbf{D}_j
\right\|_2 - \delta
\right),
\end{aligned}
\label{eq:feat_loss}
\end{equation} 
where $\lambda_{\mathrm{cos}}, \lambda_{2}, \lambda_D, \delta > 0$ are constant parameters and $\mathbf{D}^{(0)}$ is the initial dictionary before optimization. 
In \eqref{eq:feat_loss}, the first two terms correspond to standard reconstruction loss based on theline formulations that assume access to cosine similarity and L2 distance to the true (observed) embeddings $\mathbf{F}$.
The third term acts as a ``soft'' trust-region constraint \cite{nocedal1999numerical} that incurs penalty whenever the $j$th dictionary atom $\mathbf{D}_j$ deviates outside a neighborhood of its initial value $\mathbf{D}^{(0)}_j$ with radius $\delta$.
In \cref{sec:experiment:large}, we show that the proposed trust-region regularization improves performance in large-scale scenarios by mitigating overfitting. 
Although similar regularization could be applied to $\mathbf{W}$, we omit it because it has minimal impact in our experiments.

In practice, we combine the feature supervision in \eqref{eq:feat_loss} with standard photometric and depth supervision.
For photometric supervision, we use L1 and SSIM losses between the rendered RGB image $\widehat{\mathbf{I}}$ and observed image $\mathbf{I}$,
\begin{equation}
\mathcal{L}_{\mathrm{rgb}}
=
\lambda_{I_1}\,
\|\widehat{\mathbf{I}}-\mathbf{I}\|_1
+
\lambda_{I_2}\,
\mathcal{L}_{\mathrm{SSIM}}(\widehat{\mathbf{I}}, \mathbf{I}).
\label{eq:rgb_loss}
\end{equation}
For depth supervision, we compute the L1 loss on the rendered depth image $\widehat{\mathbf{Z}}$ and observed depth $\mathbf{Z}$, 
\begin{equation}
\mathcal{L}_{\mathrm{d}}
=
\|\widehat{\mathbf{Z}}-\mathbf{Z}\|_1.
\label{eq:depth_loss}
\end{equation}
Given camera pose $\mathbf{T}$,
the overall mapping problem is, 
\begin{equation}
\min_{\mathcal{G},\,\mathbf{W},\,\mathbf{D}}
\;
\lambda_{\mathrm{rgb}}\,\mathcal{L}_{\mathrm{rgb}}
+
\lambda_{\mathrm{d}}\,\mathcal{L}_{\mathrm{d}}
+
\lambda_{\mathrm{f}}\,\mathcal{L}_{\mathrm{f}},
\label{eq:full_obj}
\end{equation}
where $\lambda_{\mathrm{rgb}}$, $\lambda_{\mathrm{d}}$, and
$\lambda_{\mathrm{feat}}$ are scalar coefficients for the RGB, depth, and latent
feature loss terms, respectively.

\subsection{Online Dictionary Learning}
\label{sec:online_dictionary_construction}

We extend the dictionary-based reconstruction framework to online mapping, where RGB-D observations and their associated embeddings arrive sequentially. 
Unlike offline formulations that assume access to all measurements, 
our approach introduces a lightweight streaming initialization mechanism that provides high-quality initial dictionary atoms from incoming embeddings. 
Further, we implement efficient optimization strategies for \eqref{eq:full_obj} to satisfy the strict computational budgets required by real-time operation. 
\cref{alg:online_dict_learning} summarizes the proposed online mapping procedure upon receiving a new keyframe at time $t$.

\begin{algorithm}[t]
\caption{Online Mapping Upon Receiving Keyframe $t$}
\label{alg:online_dict_learning}
\small
\setlength{\abovedisplayskip}{2pt}
\setlength{\belowdisplayskip}{2pt}
\setlength{\abovedisplayshortskip}{1pt}
\setlength{\belowdisplayshortskip}{1pt}
\begin{algorithmic}[1]
\Require Gaussian primitives $\mathcal{G}$; previous dictionary $\mathbf{D}_{t-1}$;
current camera pose $\mathbf{T}_t$;
observed and rendered RGB image \rev{$\mathbf{I}, \widehat{\mathbf{I}}$};
observed and rendered depth \rev{$\mathbf{Z}, \widehat{\mathbf{Z}}$};
observed embeddings $\mathbf{F}_t$;
history buffer $\mathcal{H}$; history size $B$; refinement iterations $R_1, R_2$.

\vspace{0.15em}

\State \label{alg:stream_kmeans} Update dictionary via streaming K-means 
\[
\mathbf{D}_t^{(0)} \gets \mathrm{StreamKMeans}(\mathbf{F}_t,\,\mathbf{D}_{t-1},\,K)
\]

\State \label{alg:sample} Augment training data $\mathbf{F}^{\text{b}},\mathbf{T}^{\text{b}}$ with history
\[
(\mathbf{F}^{\text{b}},\,\mathbf{T}^{\text{b}})
\gets
\!(\mathbf{F}_t,\mathbf{T}_t) \, \cup \,\mathrm{Sample}(\mathcal{H},B)
\]

\State \label{alg:recon} Stage I: jointly optimize $\mathcal{G},\,\mathbf{W},\,\mathbf{D}$ for $R_1$ iterations 
\[
\mathcal{G},\,\mathbf{W},\,\mathbf{D} 
\gets
\argmin_{\mathcal{G},\,\mathbf{W},\,\mathbf{D}}
\lambda_{\mathrm{rgb}}\mathcal{L}_{\mathrm{rgb}}
+
\lambda_{\mathrm{d}}\mathcal{L}_{\mathrm{d}}
+
\lambda_{\mathrm{f}}\mathcal{L}_{\mathrm{f}}
\]
\vspace{0.1em}
\State Stage II: refine $\mathbf{W},\,\mathbf{D}$ for $R_2$ iterations \label{alg:recon2} 
\[
\mathbf{W},\,\mathbf{D} 
\gets
\argmin_{\mathbf{W},\,\mathbf{D}}
\lambda_{\mathrm{rgb}}\mathcal{L}_{\mathrm{rgb}}
+
\lambda_{\mathrm{d}}\mathcal{L}_{\mathrm{d}}
+
\lambda_{\mathrm{f}}\mathcal{L}_{\mathrm{f}}
\]

\State Update history buffer:\label{alg:history}  
$\mathcal{H}\leftarrow \mathcal{H}\cup\{(\mathbf{F}_t,\mathbf{T}_t)\}$ 

\end{algorithmic}
\end{algorithm}

\myParagraph{Dictionary Initialization from Streaming Data}
At each keyframe $t$, we first udpate dictionary atoms using a streaming K-means \cite{o2002streaming} procedure (line~\ref{alg:stream_kmeans}):
\begin{equation}
\mathbf{D}_t^{(0)} \leftarrow \mathrm{StreamKMeans}(\mathbf{F}_t,\mathbf{D}_{t-1},K),
\label{eq:stream_kmeans_init}
\end{equation}
where $\mathbf{F}_t$ denotes the newly observed embeddings, $\mathbf{D}_{t-1}$ is the current dictionary, and $K$ is the target dictionary size.
Specifically, incoming embeddings are first summarized into micro-factors (5 by default) via local K-means, and then merged and re-clustered with existing global centers.
Our design enables explicit control over dictionary size and supports fast, lightweight inference, in contrast to pairwise similarity-based methods~\cite{zou20253d, yoo2025openmonogs} that lack size control, and SVD-based approaches~\cite{zhang2021invertible} that are computationally expensive. 
Empirically, we observe that the dictionary size $K$ and micro-cluster size have little effect on performance, because the input semantic features are already well structured by upstream grouping from SAM masks or LSeg features. 
All steps are implemented using batched computation for GPU, allowing real-time clustering performance. 

\myParagraph{Two-stage Optimization}
Unlike offline methods that allow sufficient optimization time, jointly optimizing dictionary and Gaussian parameters online is time consuming. 
Therefore, we optimize \eqref{eq:full_obj} in two stages. 
In stage I, we carry out joint optimization of the Gaussian primitives $\mathcal{G}$, the projection matrix $\mathbf{W}$, and the dictionary $\mathbf{D}$ for a small number of $R_1$ iterations (1 by default); see line~\ref{alg:recon}.
In Stage II, we perform a lightweight refinement that further optimizes $\mathbf{W}$ and $\mathbf{D}$ to improve feature reconstruction fidelity within a limited optimization budget, for $R_2$ (5 by default) iterations while keeping $\mathcal{G}$ fixed; see line~\ref{alg:recon2}.

To address the catastrophic forgetting problem, we introduce a historical learning strategy (in lines~\ref{alg:recon}, \ref{alg:recon2}, \ref{alg:history}) similar to prior work~\cite{pan2024pin}. 
At each time step, we randomly sample $B$ past keyframes from the history buffer to augment the newly arrived keyframe,  enabling the model to retain previous information while adapting to new observations. 
The history size $B$ controls the tradeoff between reducing catastrophic forgetting and maintaining computational efficiency. 
Together with the trust-region regularization in Eq.~\eqref{eq:feat_loss}, this strategy effectively alleviates catastrophic forgetting during online mapping. 

\begin{figure*}[t]
	\centering
	\includegraphics[width=.95\textwidth]{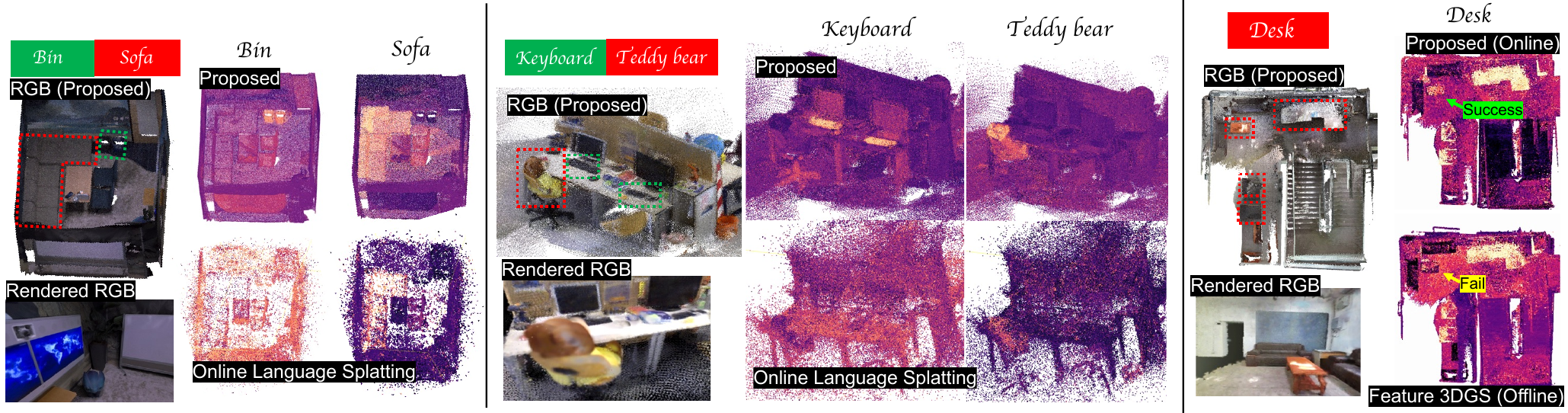}
	\caption{3D segmentation results of the proposed method and baselines on Office0 (left), TUM3 (middle), and Floor2 (right) scenes. In the Floor2 scene, Feature-3DGS fails to detect the desk segments (yellow arrow) while our method succeeds (green arrow).} 
	\label{fig:qualitative_result_3d}
	\vspace{-2mm}
\end{figure*}

\subsection{Large-Scale Mapping via Voxel Hashing}
\label{sec:large_scale_mapping}
\myParagraph{System Overview} 
Building on the proposed algorithms, we present an online RGB-D 3DGS mapping system that incrementally reconstructs large-scale scenes while jointly modeling geometry, appearance, and latent embeddings, and maintains bounded GPU memory usage by storing the global Gaussian map on the CPU and optimizing only an active local map on the GPU.

First, camera poses are tracked using an external real-time tracker~\cite{ha2024rgbd, campos2021orb}.
Depending on the tracker, 
keyframes are selected based on the map–scan overlap for~\cite{ha2024rgbd} and uniform interval sampling for~\cite{campos2021orb}. For each selected keyframe, newly initialized 3D Gaussians are inserted into a voxel-hashed global map stored in CPU memory, discussed in full detail in the next paragraph.
At any time, the system maintains a local map around the robot by retrieving Gaussians within a specific radius around the current camera position from the global map and transferring them to GPU memory. 
Then, these local Gaussians are optimized using photometric, depth, and feature losses, as discussed in Sec. \ref{sec:feature_splatting}-\ref{sec:online_dictionary_construction}. 

\myParagraph{Local-Global Map Management via Voxel Hashing} 
To avoid duplicate primitives and enable fast spatial indexing, we organize the global map using voxel-hashing.
Each Gaussian is assigned to a voxel $\mathbf{v}_i=\lfloor \mathbf{p}_i / s_v \rfloor$, where $s_v$ denotes the voxel size (0.01~m by default).
A hash table $h:\mathbb{Z}^3\rightarrow\mathbb{N}$ maps integer voxel coordinates to their corresponding indices in CPU memory, storing at most one Gaussian per voxel. 
Hash collisions may occur when distinct voxel coordinates map to the same index, but they are rare and we observe that they have negligible impact on performance. 
Furthermore, since each rendered pixel aggregates contributions from many 3D Gaussians, alpha blending \eqref{eq:alpha_blending} makes the influence of a single collided primitive insignificant.

The global and local maps are synchronized periodically when the traveled distance since the last synchronization exceeds a predefined threshold. 
Synchronization has two steps.
In the \emph{Global$\rightarrow$Local} step, the current local map is pruned to retain only Gaussians within the active radius, after which additional nearby Gaussians are retrieved from the voxel hash table and appended without duplication. 
In the \emph{Local$\rightarrow$Global} step, locally optimized Gaussians are inserted back to the global map. 
Gaussians corresponding to existing global primitives, identified via voxel-hash lookup, directly update their associated entries, while newly created Gaussians are inserted by computing their voxel-hash values only if the voxel is unoccupied. 
Our implementation does not rely on any specialized CPU/GPU parallelization beyond standard PyTorch operations. 

\begin{table*}[h]
\centering
\caption{Quantitative results on TUM~\cite{sturm2012benchmark}, Replica~\cite{straub2019replica}, and FastCaMo-Large~\cite{tang2023mips}. The best and second-best are shown in green and yellow. OOM denotes out-of-memory. OmniMap does not report cosine loss since it does not map VLM embeddings. }
\label{tab:all_scenes_tum_replica}
\resizebox{0.90\textwidth}{!}{
\small
\begin{tabular}{l c ccccc ccccccccc ccccc}
\toprule
{\textbf{Methods}} & {\textbf{Metric}}
& \multicolumn{3}{c}{\textbf{TUM}}
& \multicolumn{8}{c}{\textbf{Replica}} & \multicolumn{5}{c}{\textbf{FastCaMo-Large}}\\ 
\cmidrule(lr){3-5}\cmidrule(lr){6-13}\cmidrule(lr){14-18}
& &
TUM1 & TUM2 & TUM3
& Room0 & Room1 & Room2
& Office0 & Office1 & Office2 & Office3 & Office4 
& Lab & Floor1 & Floor2 & Stair1 & Stair2 \\
\midrule

\multirow{4}{*}{Feature3DGS~\cite{zhou2024feature}}
& Cos-loss $\downarrow$
& \second{0.130} & \best{0.078} & \second{0.100}
& \second{0.078} & \second{0.072} & \second{0.077}
& \second{0.096} & \second{0.058} & \second{0.065} & \second{0.078} & \second{0.064}
& \second{0.119} & \best{0.045} & \best{0.074} & \best{0.043} & \best{0.033} \\
& mIoU $\uparrow$
& 0.576 & 0.949 & 0.797
& 0.635 & 0.543 & 0.345
& 0.447 & 0.383 & 0.532 & 0.133 & 0.413
& 0.344 & \second{0.637} & \second{0.493} & 0.223 & 0.224 \\
& PSNR $\uparrow$
& 15.00 & 19.49 & 15.59
& \second{22.63} & 25.52 & 24.28
& \best{38.04} & \second{38.02} & 22.64 & 24.93 & 23.95
& 15.95 & 22.54 & \best{24.21} & \second{23.71} & 17.29 \\
& FPS $\uparrow$
& 0.77 & 0.67 & 0.67
& 0.54 & 0.65 & 0.63
& 0.61 & 0.68 & 0.62 & 0.58 & 0.62
& 0.74 & 0.78 & 0.79 & 0.92 & 0.93 \\
\midrule

\multirow{4}{*}{M3~\cite{zou20253d}}
& Cos-loss $\downarrow$
& 0.136 & 0.116 & 0.133
& 0.091 & 0.105 & 0.108
& 0.141 & 0.114 & 0.119 & 0.132 & 0.123
& 0.121 & OOM & OOM & 0.074 & 0.081 \\
& mIoU $\uparrow$
& 0.674 & \second{0.957} & \best{0.910}
& \best{0.883} & \second{0.845} & \second{0.752}
& \second{0.819} & 0.414 & 0.462 & 0.308 & 0.422
& \second{0.588} & OOM & OOM & \best{0.507} & \second{0.472} \\
& PSNR $\uparrow$
& 13.49 & 19.10 & 16.22
& 21.62 & 25.52 & 24.51
& \second{38.04} & 38.00 & 22.69 & 25.20 & \second{24.46}
& 16.79 & OOM & OOM & 19.70 & 17.38 \\
& FPS $\uparrow$
& 0.77 & 0.67 & 0.67
& 0.53 & 0.65 & 0.63
& 0.61 & 0.68 & 0.62 & 0.57 & 0.62
& 0.73 & OOM & OOM & 0.90 & 0.93 \\
\midrule

\multirow{3}{*}{OmniMap~\cite{deng2025omnimap}}
& mIoU $\uparrow$
& 0.449 & 0.405 & 0.391
& 0.492 & 0.665 & 0.590
& 0.282 & 0.207 & 0.415 & 0.522 & 0.419
& 0.149 & 0.189 & 0.292 & 0.337 & 0.453 \\
& PSNR $\uparrow$
& \best{19.21} & \best{20.23} & \best{21.25}
& 22.08 & 23.25 & 21.95
& 26.17 & 27.00 & 20.39 & 21.77 & 23.13
& \second{16.88} & \second{23.27} & 22.45 & 23.58 & \best{22.76} \\
& FPS $\uparrow$
& \second{6.98} & \second{7.80} & \second{7.97}
& \second{7.34} & \second{8.63} & \second{8.12}
& \second{8.87} & \second{9.10} & \second{7.26} & \second{6.84} & \second{7.22}
& \second{6.39} & \second{6.33} & \second{6.46} & \second{7.13} & \second{7.80} \\
\midrule

\multirow{4}{*}{OnlineLangSplat~\cite{katragadda2025online}}
& Cos-loss $\downarrow$
& 0.280 & 0.275 & 0.274
& 0.222 & 0.269 & 0.232
& 0.290 & 0.206 & 0.206 & 0.218 & 0.234
& 0.314 & 0.287 & 0.296 & 0.270 & 0.264 \\
& mIoU $\uparrow$
& \second{0.714} & 0.588 & 0.697
& 0.609 & 0.321 & 0.484
& 0.774 & \second{0.934} & \best{0.884} & \best{0.709} & \best{0.692}
& 0.504 & 0.522 & 0.425 & 0.241 & 0.247 \\
& PSNR $\uparrow$
& \second{18.02} & 15.58 & 18.74
& \best{29.77} & \best{34.44} & \best{31.65}
& 37.04 & 37.13 & \best{31.46} & \best{34.85} & \best{33.60}
& 11.83 & 12.27 & 11.62 & 12.00 & 13.86 \\
& FPS $\uparrow$
& 0.43 & 0.44 & 0.37
& 1.13 & 1.08 & 1.23
& 1.38 & 1.49 & 1.37 & 1.38 & 1.12
& 1.52 & 1.44 & 1.57 & 1.51 & 1.55 \\
\midrule

\multirow{4}{*}{\shortstack[l]{OnlineLangSplat~\cite{katragadda2025online} \\ 
+ GS-ICP~\cite{ha2024rgbd}}}
& Cos-loss $\downarrow$
& 0.293 & 0.315 & 0.288
& 0.258 & 0.244 & 0.240
& 0.324 & 0.244 & 0.257 & 0.258 & 0.260
& 0.326 & 0.292 & 0.319 & 0.259 & 0.260 \\
& mIoU $\uparrow$
& 0.688 & 0.490 & 0.598
& 0.545 & 0.523 & 0.463
& 0.548 & 0.767 & 0.673 & 0.535 & \second{0.620}
& 0.488 & 0.493 & 0.355 & 0.276 & 0.252 \\
& PSNR $\uparrow$
& 16.32 & 16.32 & \second{18.80}
& 21.01 & 18.38 & 21.28
& 19.32 & 25.92 & 20.56 & 20.99 & 19.80
& 10.89 & 11.52 & 10.27 & 14.60 & 14.94 \\
& FPS $\uparrow$
& 2.50 & 2.21 & 2.33
& 3.44 & 2.93 & 3.58
& 2.76 & 3.62 & 3.20 & 3.55 & 2.86
& 2.31 & 2.63 & 2.68 & 2.30 & 1.91 \\
\midrule

\multirow{4}{*}{LatentAM (ours)}
& Cos-loss $\downarrow$
& \best{0.110} & \second{0.096} & \best{0.094}
& \best{0.069} & \best{0.055} & \best{0.061}
& \best{0.083} & \best{0.045} & \best{0.058} & \best{0.068} & \best{0.056}
& \best{0.090} & \second{0.079} & \second{0.074} & \second{0.043} & \second{0.042} \\
& mIoU $\uparrow$
& \best{0.792} & \best{0.960} & \second{0.885}
& \second{0.686} & \best{0.873} & \best{0.844}
& \best{0.874} & \best{0.968} & \second{0.687} & \second{0.674} & 0.453
& \best{0.758} & \best{0.668} & \best{0.613} & \second{0.457} & \best{0.476} \\
& PSNR $\uparrow$
& 15.09 & \second{19.50} & 15.46
& 19.97 & \second{25.57} & \second{26.35}
& 38.00 & \best{38.06} & \second{22.77} & \second{26.95} & 23.11
& \best{19.67} & \best{25.09} & \second{23.25} & \best{23.93} & \second{22.31} \\
& FPS $\uparrow$
& \best{20.24} & \best{20.17} & \best{17.89}
& \best{18.63} & \best{17.90} & \best{14.78}
& \best{19.82} & \best{22.40} & \best{20.42} & \best{17.41} & \best{18.86}
& \best{13.64} & \best{17.07} & \best{15.30} & \best{28.62} & \best{25.34} \\
\bottomrule
\end{tabular}}
\vspace{0mm}
\end{table*}

\section{Experiments}
\label{sec:experiment}

We comprehensively evaluate \name, showing state-of-the-art feature reconstruction fidelity and significantly faster runtime than existing methods. 
Furthermore, we demonstrate scalability of our approach on a large-scale multi-floor custom dataset. 
Lastly, we ablate the contributions of key components, including dictionary learning, streaming K-means–based dictionary construction, and local map management.

\subsection{Experimental Setup}
We conduct experiments on three public datasets: Replica~\cite{straub2019replica}, TUM~\cite{sturm2012benchmark}, and the large-scale FastCaMo dataset~\cite{tang2023mips}. 
To evaluate scalability and robustness in large-scale, long-term environments, we also capture a custom dataset using a Gemini 335Le RGB-D camera (\cref{sec:experiment:large}). 
For the hyperparameters of the proposed method, we use $\delta = 0.1$, $\lambda_{I_1} = 0.2$, $\lambda_{I_2} = 0.8$, $\lambda_d = 0.1$, $\lambda_f = 5.0$, $\lambda_{\text{cos}} = 0.5$, $\lambda_2 = 0.5$, and $\lambda_D = 0.2$ with the learning rates for $\mathbf{W}$ and $\mathbf{D}$ set to $0.01$. 
\rev{Moreover, we empirically set $B=5$ and $K=2000$ to balance feature retention, dictionary capacity, and runtime across different scene scales.} 

\myParagraph{Baselines} 
We select state-of-the-art offline feature splatting approaches, including Feature-3DGS~\cite{zhou2024feature} and M3~\cite{zou20253d}, as well as state-of-the-art online methods, including Online Language Splatting~\cite{katragadda2025online} for feature splatting SLAM and OmniMap~\cite{deng2025omnimap} for codebook-based semantic mapping. 
While OpenMonoGS-SLAM~\cite{yoo2025openmonogs} is also related, 
the code is not available so we analyze its core idea through an ablation in \cref{sec:experiment:ablation}. 
For offline baselines and OmniMap, we use the same pose tracker~\cite{ha2024rgbd} and keyframe selection as ours for fair comparison. 
Each offline model is trained for 20 epochs, which is sufficient for full convergence across all scenes. We set the query dimension $d_s=32$ for both the offline methods and our method. 
\rev{For Online Language Splatting~\cite{katragadda2025online}, we follow the original two-stage training pipeline, in which both the autoencoder and Gaussian parameters are optimized using a pretrained super-resolution decoder, and report results with both the original MonoGS~\cite{matsuki2024gaussian} tracker and GS-ICP~\cite{ha2024rgbd} tracker to provide a fair and controlled runtime comparison using the same MonoGS keyframes.}

\myParagraph{Performance Metrics}
We use the cosine loss, defined as one minus the cosine similarity between the reconstructed and ground-truth 2D VLM embeddings. 
Following \cite{qiu2024feature,zhou2024feature,katragadda2025online}, we also compute the mean Intersection-over-Union (mIoU) over a predefined set of 10 semantic query words. 
Per-class semantic probabilities are computed by inner products between 2D embeddings and text embeddings, followed by softmax. 
For OmniMap, which performs instance segmentation instead of feature mapping, each 3D instance is projected to the 2D plane and re-grouped by query words to directly compute mIoU. 
For photometric evaluation, we compute the peak signal to noise ratio (PSNR) on the rendered RGB images. 
Although our main evaluation uses CLIP~\cite{radford2021learning} and SAM~\cite{kirillov2023segment}, \name is model-agnostic and can operate with a variety of VLMs. 
To demonstrate this feature, we also present evaluation using DINOv3~\cite{simeoni2025dinov3} and LSeg~\cite{li2022language}. 
Finally, we report frame per second (FPS) to evaluate end-to-end runtime including VLM feature extraction. 
Experiments are conducted on a workstation with NVIDIA RTX~5090 GPU, 64~GB RAM, and Intel Core Ultra 9 285K CPU. 

\subsection{Evaluation of Mapping Performance} 
Table~\ref{tab:all_scenes_tum_replica} reports the quantitative results on TUM, Replica, and FastCaMo datasets.
Qualitative 3D open-set segmentation results are shown in Fig.~\ref{fig:qualitative_result_3d}.

On TUM and Replica, our method consistently outperforms the compared methods in both feature reconstruction and open-vocabulary language tasks, while maintaining real-time performance. In particular, our method outperforms both offline distillation-based and dictionary-based approaches, thanks to the proposed dictionary learning that is designed to effectively adapt to diverse scene variations. 
Moreover, OmniMap relies on ground-truth poses in its original benchmark~\cite{deng2025omnimap} and performs instance-level mapping; replacing the poses with GS-ICP~\cite{ha2024rgbd} leads to unstable instance tracking, whereas our feature–based mapping remains robust. 
On FastCaMo-Large, our method achieves state-of-the-art performance.
Although Feature-3DGS is better on some metrics, it is offline and requires an environment- and model-dependent decoder.
In contrast, ours is online, pretraining-free, and VLM-agnostic.
M3 runs out of memory in large-scale scenes due to unbounded dictionary growth, whereas our streaming K-means keeps the dictionary size fixed.

Although our photometric rendering quality, measured by PSNR in Table~\ref{tab:all_scenes_tum_replica}, is lower than that of Online Language Splatting, this difference mainly stems from the number of optimization iterations. Our method uses only one Stage I iteration and five Stage II iterations to prioritize real-time operation, whereas Online Language Splatting uses 150 iterations by default, improving photometric quality at a substantially higher computational cost. Moreover, replacing MonoGS with the faster GS-ICP tracker~\cite{ha2024rgbd} for Online Language Splatting yields only modest FPS gains while substantially degrading reconstruction quality. This is because MonoGS supports sparse keyframes by aligning frames using rendered photometric and depth information, whereas GS-ICP relies only on geometric alignment and therefore requires denser keyframes for stable tracking. 
As a result, GS-ICP with Online Language Splatting under the original sparse keyframe setting suffers from drift.

\begin{table}[t]
\centering
\caption{Comparison under different visual foundation models on the TUM dataset.
Best results are highlighted in green.}
\label{tab:vfm_ablation}
\renewcommand{\arraystretch}{0.85}
\setlength{\tabcolsep}{4pt}
\resizebox{0.99\columnwidth}{!}{
\begin{tabular}{c l ccc ccc ccc}
\toprule
& & \multicolumn{3}{c}{\textbf{CLIP-SAM}}
& \multicolumn{3}{c}{\textbf{DINOv3-SAM}}
& \multicolumn{3}{c}{\textbf{LSeg}} \\
\cmidrule(lr){3-5}\cmidrule(lr){6-8}\cmidrule(lr){9-11}
\textbf{Seq.} & \textbf{Method}
& $\mathcal{L}_{\cos}\!\downarrow$ & $\mathcal{L}_{2}\!\downarrow$ & FPS$\uparrow$
& $\mathcal{L}_{\cos}\!\downarrow$ & $\mathcal{L}_{2}\!\downarrow$ & FPS$\uparrow$
& $\mathcal{L}_{\cos}\!\downarrow$ & $\mathcal{L}_{2}\!\downarrow$ & FPS$\uparrow$ \\
\midrule

\multirow{3}{*}{TUM1}
& Feat3DGS~\cite{zhou2024feature}
& 0.130 & 0.077 & 0.77
& 0.231 & 0.029 & 0.57
& 0.069 & 0.00028 & 0.56 \\
& M3~\cite{zou20253d}
& 0.136 & 0.100 & 0.77
& 0.192 & 0.024 & 0.54
& 0.134 & 0.00051 & 0.55 \\
& LatentAM (ours)
& \best{0.110} & \best{0.073} & \best{20.24}
& \best{0.185} & \best{0.020} & \best{16.99}
& \best{0.064} & \best{0.00024} & \best{17.29} \\
\midrule

\multirow{3}{*}{TUM2}
& Feat3DGS~\cite{zhou2024feature}
& \best{0.078} & \best{0.052} & 0.67
& \best{0.103} & \best{0.013} & 0.52
& \best{0.035} & 0.00022 & 0.55 \\
& M3~\cite{zou20253d}
& 0.116 & 0.121 & 0.67
& 0.147 & 0.021 & 0.80
& 0.132 & 0.00052 & 0.55 \\
& LatentAM (ours)
& 0.096 & 0.061 & \best{20.17}
& 0.145 & 0.014 & \best{23.11}
& 0.053 & \best{0.00020} & \best{30.57} \\
\midrule

\multirow{3}{*}{TUM3}
& Feat3DGS~\cite{zhou2024feature}
& 0.100 & 0.119 & 0.67
& 0.197 & 0.021 & 0.48
& \best{0.049} & 0.00029 & 0.51 \\
& M3~\cite{zou20253d}
& 0.133 & 0.132 & 0.67
& 0.198 & 0.020 & 0.50
& 0.119 & 0.00044 & 0.54 \\
& LatentAM (ours)
& \best{0.094} & \best{0.060} & \best{17.89}
& \best{0.192} & \best{0.017} & \best{20.60}
& 0.068 & \best{0.00026} & \best{26.40} \\
\bottomrule
\end{tabular}}
\renewcommand{\arraystretch}{1.0}
\end{table}

\begin{figure}[t]
    \centering
    \includegraphics[width=1.0\linewidth]{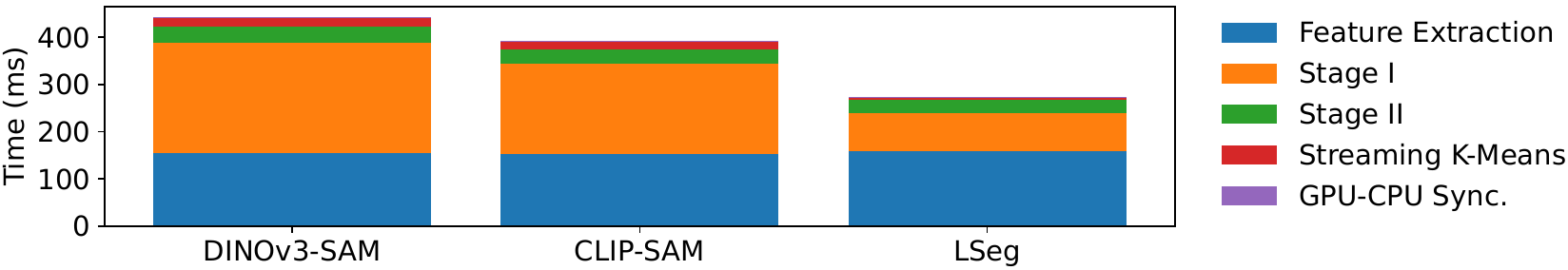}
    \caption{Breakdown of mapping runtime (ms) per keyframe using different VLM backbones on the TUM datasets.}
    \label{fig:break_down_timefig1}
    \vspace{0mm}
\end{figure}

To demonstrate the model-agnostic mapping capability of \name, we evaluate on alternative VLMs including DINOv3~\cite{simeoni2025dinov3} and LSeg~\cite{li2022language} in Table~\ref{tab:vfm_ablation}. 
The proposed method outperforms other offline methods on TUM1 and TUM3 across all VLM settings. 
In TUM2, the scene is highly repetitive and exhibits minimal camera motion. As a result, the distillation-based method, Feature-3DGS, achieves slightly better performance, although our method still demonstrates comparable feature splatting performance to Feature-3DGS. 

Note that our method achieves 12 FPS with short keyframe intervals (every 4 frames on average), and over 30 FPS with longer intervals (every 10 frames on average). In practice, the keyframe intervals are determined adaptively by the tracker.~\cite{ha2024rgbd}. 
To provide more insights, Fig.~\ref{fig:break_down_timefig1} reports the per-keyframe runtime breakdown under different VLM choices, showing that the mapping runtime remains under $450$~ms per keyframe, including VLM feature extraction.

\begin{figure}[t]
	\centering
	\includegraphics[width=.99\columnwidth]{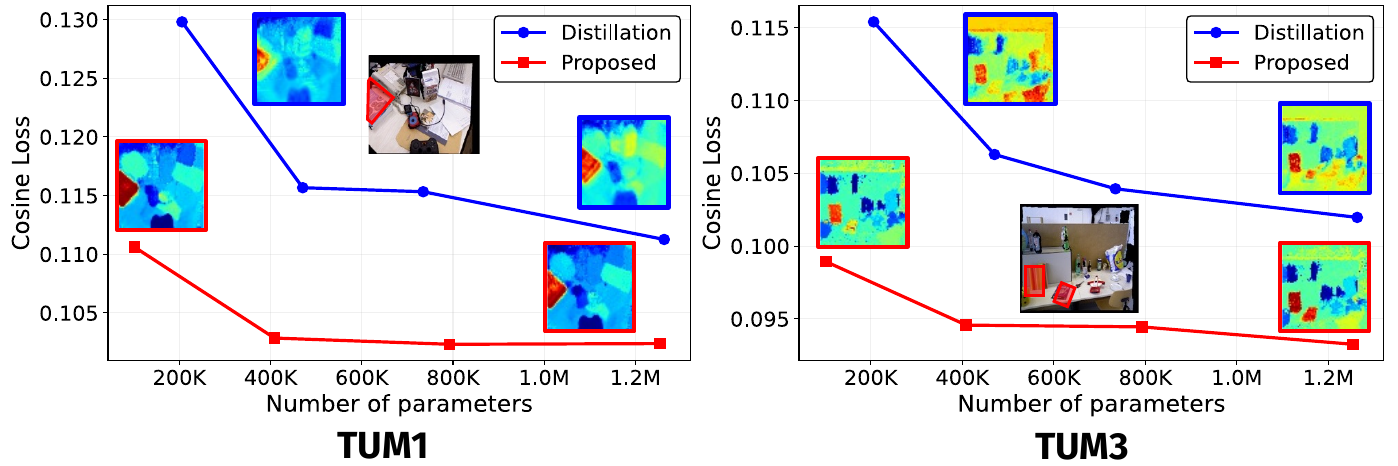}
	\caption{Comparison between dictionary learning and distillation \cite{zhou2024feature} on TUM1 and TUM3. The semantic segmentation results are overlaid, with the ground-truth object (book) highlighted. }
	\label{fig:learnable_params}
	\vspace{0mm}
\end{figure}

\begin{table}[t]
\centering
\caption{Ablation study on online dictionary learning. K-means denotes streaming K-means–based dictionary construction, and DL denotes dictionary learning. Best results are highlighted in green.}
\label{tab:ablation_dict}
\resizebox{0.77\columnwidth}{!}{
\begin{tabular}{c c c cc cc cc}
\toprule
\multirow{2}{*}{K-means} & \multirow{2}{*}{DL} 
& \multicolumn{2}{c}{TUM1} & \multicolumn{2}{c}{TUM2} & \multicolumn{2}{c}{TUM3} \\
& & Cos $\downarrow$ & mIoU $\uparrow$
  & Cos $\downarrow$ & mIoU $\uparrow$
  & Cos $\downarrow$ & mIoU $\uparrow$ \\
\midrule
$\times$  & $\times$ 
& 0.146 & 0.506
& 0.137 & 0.546
& 0.124 & 0.602 \\
$\checkmark$ & $\times$
& 0.127 & 0.529
& 0.104 & 0.780
& 0.111 & 0.593 \\
$\times$  & $\checkmark$ 
& 0.141 & 0.647
& 0.124 & 0.766
& 0.124 & 0.765 \\
$\checkmark$ & $\checkmark$
& \best{0.110} & \best{0.792}
& \best{0.096} & \best{0.960}
& \best{0.094} & \best{0.885} \\
\bottomrule
\end{tabular}}
\vspace{-2mm}
\end{table}

\begin{table}[t]
\centering
\caption{Peak GPU usage (MB) during mapping.
The best and second-best results are highlighted in green and yellow, respectively.}
\label{tab:ablation_local}
\resizebox{0.8\columnwidth}{!}{
\begin{tabular}{l cc cc cc}
\toprule
& \multicolumn{2}{c}{TUM1} & \multicolumn{2}{c}{TUM2} & \multicolumn{2}{c}{TUM3} \\
Method
& Cos $\downarrow$ & Mem. $\downarrow$
& Cos $\downarrow$ & Mem. $\downarrow$
& Cos $\downarrow$ & Mem. $\downarrow$ \\
\midrule

M3~\cite{zou20253d}
& 0.136 & \second{5188.40}
& 0.116 & \second{12852.8}
& 0.133 & \second{10767.2} \\

LatentAM (w/o local)
& \best{0.105} & 6428.77
& \best{0.087} & 14133.2
& \best{0.091} & 13845.0 \\

LatentAM (with local)
& \second{0.110} & \best{5180.16}
& \second{0.096} & \best{7388.47}
& \second{0.094} & \best{8538.66} \\
\bottomrule
\end{tabular}}
\vspace{0mm}
\end{table}

\begin{figure}[t]
    \centering
    \includegraphics[width=0.8\linewidth]{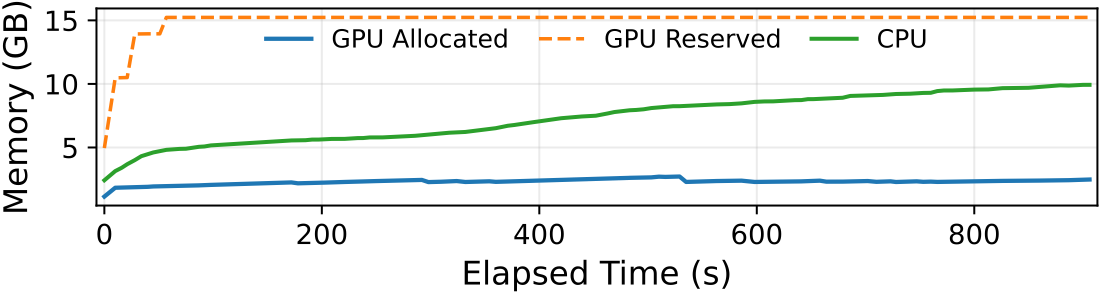}
    \caption{CPU/GPU memory over time on the large-scale dataset.}
    \label{fig:memory-plot2}
    \vspace{0mm}
\end{figure}

\begin{figure}[!t]
	\centering
    \includegraphics[width=.9\columnwidth]{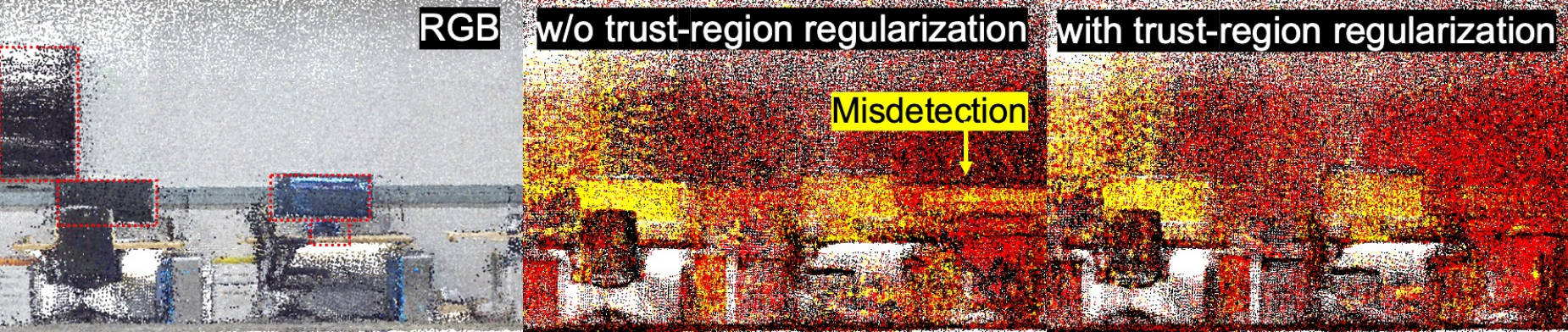}
	\caption{
    Comparison of segmentation results with and without trust-region regularization. The target word is \emph{monitor}. Without regularization, the model misdetects monitors on the wall. 
    }
	\label{fig:regulation}
	\vspace{-2mm}
\end{figure}

\subsection{Ablation Study}
\label{sec:experiment:ablation}
\myParagraph{Dictionary Learning vs. Distillation}
To demonstrate the efficiency of our online dictionary learning approach, we present a systematic evaluation compared to decoder-based distillation methods such as Feature-3DGS~\cite{zhou2024feature}.
We report performance with respect to the number of learnable parameters excluding Gaussian primitives. 
The number of learnable parameters is controlled by the dictionary size in our method and by the decoder depth in the distillation-based method. 
For Feature-3DGS, we add skip connections to the decoder to mitigate vanishing gradients in deeper layers. 

As shown in Fig.~\ref{fig:learnable_params}, distillation-based methods remain consistently weaker than ours despite operating in an offline setting. 
This observation agrees with prior work \cite{zou20253d} and suggests that pure distillation introduces distortion between the recovered and original embeddings. 
In contrast, our method maintains a compact dictionary of latent embeddings from past observations as explicit memory, which enables higher fidelity reconstruction. Moreover, unlike decoder-based approaches whose performance strongly depends on pretraining data selection, our dictionary is learned directly from streaming data during online mapping.

\myParagraph{Online Dictionary Construction} 
We evaluate four variants of the proposed online dictionary learning method: 
(i) \emph{Random + w/o DL}, where the dictionary is randomly initialized from past keyframes and kept static; 
(ii) \emph{Random + with DL}, where the randomly initialized dictionary is optimized online; 
(iii) \emph{Kmeans + w/o DL}, where the dictionary is constructed by streaming K-means and kept static; 
and (iv) \emph{Kmeans + with DL}, our full approach combining streaming K-means initialization with dictionary optimization.
Quantitative results are reported in Table~\ref{tab:ablation_dict}. 
Overall, both streaming K-means and online dictionary learning improve performance.
Notably, refining the dictionary via the proposed optimization yields up to 59.4\% improvement on the mIoU metric, which clearly demonstrate its advantage compared to static dictionary used by related work~\cite{yoo2025openmonogs}.

\myParagraph{Local Mapping}
Lastly, we evaluate the efficiency of our local mapping strategy using peak GPU memory.
As shown in Table~\ref{tab:ablation_local}, local mapping substantially reduces memory compared to M3 and our method without local mapping. 
 
\subsection{Large Scale Mapping}
\label{sec:experiment:large}
We collect a large-scale, long-term dataset in a campus building. The dataset spans $14.6$~minutes of continuous operation across two floors connected by a staircase, including corridor traversal and four room transitions, with a total traveled distance of approximately $530$~m, as shown in Fig.~\ref{fig:mot}. 
For this experiment, we use ORB-SLAM3~\cite{campos2021orb} with loop-closure disabled as the pose tracker due to its stability in large-scale environments. 
Our method successfully reconstructs both 3D photometric and latent representations over the entire trajectory. 
Using CLIP embeddings, the reconstructed map directly supports open-vocabulary semantic queries at large scale.
In contrast, all baselines, including M3, Feature3DGS, and Online Language Splatting, fail to complete mapping due to GPU memory exhaustion. 
Fig.~\ref{fig:memory-plot2} reports the CPU and GPU memory usage over time. 
Thanks to the proposed local-global map management strategy, our method significantly improves memory efficiency and maintains bounded GPU memory. 
Fig.~\ref{fig:regulation} qualitatively evaluates the constrained dictionary learning formulation in \eqref{eq:feat_loss}, where trust-region regularization improves segmentation results and reconstruction quality in large scenes.

\section{Conclusions}
We presented \textbf{\name}, an online 3DGS-based mapping system that reconstructs geometry, appearance, and VLM embeddings in real time for scalable and versatile 3D perception tasks. 
By formulating feature splatting as online dictionary learning with streaming K-means initialization and trust-region regularization, \name achieves high-fidelity embedding reconstruction without pretraining or model-specific distillation. Our local mapping strategy further enables memory-efficient long-term, large-scale mapping. 
Future work will focus on improving photometric rendering quality, adaptive dictionary-size management, and improving robustness under severe domain shifts.

\bibliographystyle{IEEEtran}
\bibliography{ICRA2024}

@book{nocedal1999numerical,
  title={Numerical optimization},
  author={Nocedal, Jorge and Wright, Stephen J},
  year={1999},
  publisher={Springer}
}

@inproceedings{zou20253d,
  title     = {{3D}-{Spatial} Multimodal Memory},
  author    = {Zou, Xueyan and Song, Yuchen and Qiu, Ri-Zhao and Peng, Xuanbin and Ye, Jianglong and Liu, Sifei and Wang, Xiaolong},
  booktitle = {Proc. Int. Conf. Learn. Represent. (ICLR)},
  year      = {2025}
}

@inproceedings{zhou2024feature,
  title     = {Feature {3DGS}: Supercharging {3D} Gaussian Splatting to Enable Distilled Feature Fields},
  author    = {Zhou, Shijie and Chang, Haoran and Jiang, Sicheng and Fan, Zhiwen and Zhu, Zehao and Xu, Dejia and Chari, Pradyumna and You, Suya and Wang, Zhangyang and Kadambi, Achuta},
  booktitle = {Proc. IEEE/CVF Conf. Comput. Vis. Pattern Recognit. (CVPR)},
  year      = {2024}
}

@inproceedings{qiu2024feature,
  title     = {Language-Driven Physics-Based Scene Synthesis and Editing via Feature Splatting},
  author    = {Qiu, Ri-Zhao and Yang, Ge and Zeng, Weijia and Wang, Xiaolong},
  booktitle = {Proc. Eur. Conf. Comput. Vis. (ECCV)},
  year      = {2024}
}

@inproceedings{radford2021learning,
  title     = {Learning Transferable Visual Models from Natural Language Supervision},
  author    = {Radford, Alec and Kim, Jong Wook and Hallacy, Chris and Ramesh, Aditya and Goh, Gabriel and Agarwal, Sandhini and Sastry, Girish and Askell, Amanda and Mishkin, Pamela and Clark, Jack and others},
  booktitle = {Proc. Int. Conf. Mach. Learn. (ICML)},
  year      = {2021}
}

@article{simeoni2025dinov3,
  title   = {{DINOv3}},
  author  = {Sim{\'e}oni, Oriane and Vo, Huy V. and Seitzer, Maximilian and Baldassarre, Federico and Oquab, Maxime and Jose, Cijo and Khalidov, Vasil and Szafraniec, Marc and Yi, Seungeun and Ramamonjisoa, Micha{\"e}l and others},
  journal = {arXiv preprint arXiv:2508.10104},
  year    = {2025}
}

@inproceedings{kirillov2023segment,
  title     = {Segment Anything},
  author    = {Kirillov, Alexander and Mintun, Eric and Ravi, Nikhila and Mao, Hanzi and Rolland, Chloe and Gustafson, Laura and Xiao, Tete and Whitehead, Spencer and Berg, Alexander C. and Lo, Wan-Yen and others},
  booktitle = {Proc. IEEE/CVF Int. Conf. Comput. Vis. (ICCV)},
  year      = {2023}
}

@inproceedings{li2022language,
  title     = {Language-Driven Semantic Segmentation},
  author    = {Li, Boyi and Weinberger, Kilian Q. and Belongie, Serge and Koltun, Vladlen and Ranftl, Ren{\'e}},
  booktitle = {Proc. Int. Conf. Learn. Represent. (ICLR)},
  year      = {2022}
}

@inproceedings{katragadda2025online,
  title     = {Online Language Splatting},
  author    = {Katragadda, Saimouli and Wu, Cho-Ying and Guo, Yuliang and Huang, Xinyu and Huang, Guoquan and Ren, Liu},
  booktitle = {Proc. IEEE/CVF Int. Conf. Comput. Vis. (ICCV)},
  year      = {2025}
}

@article{pan2024pin,
  title     = {{PIN}-{SLAM}: {LiDAR} {SLAM} Using a Point-Based Implicit Neural Representation for Achieving Global Map Consistency},
  author    = {Pan, Yue and Zhong, Xingguang and Wiesmann, Louis and Posewsky, Thorbj{\"o}rn and Behley, Jens and Stachniss, Cyrill},
  journal   = {IEEE Trans. Robot.},
  year      = {2024},
  publisher = {IEEE}
}

@inproceedings{wu2024opengaussian,
  title   = {OpenGaussian: Towards Point-Level {3D} Gaussian-Based Open-Vocabulary Understanding},
  author  = {Wu, Yanmin and Meng, Jiarui and Li, Haijie and Wu, Chenming and Shi, Yahao and Cheng, Xinhua and Zhao, Chen and Feng, Haocheng and Ding, Errui and Wang, Jingdong and others},
  booktitle = {Proc. Adv. Neural Inf. Process. Syst. (NeurIPS)},
  year    = {2024}
}

@inproceedings{kerr2023lerf,
  title     = {{LERF}: Language Embedded Radiance Fields},
  author    = {Kerr, Justin and Kim, Chung Min and Goldberg, Ken and Kanazawa, Angjoo and Tancik, Matthew},
  booktitle = {Proc. IEEE/CVF Int. Conf. Comput. Vis. (ICCV)},
  year      = {2023}
}

@inproceedings{yang2025opengs,
  title     = {OpenGS-{SLAM}: Open-Set Dense Semantic {SLAM} with {3D} Gaussian Splatting for Object-Level Scene Understanding},
  author    = {Yang, Dianyi and Gao, Yu and Wang, Xihan and Yue, Yufeng and Yang, Yi and Fu, Mengyin},
  booktitle = {Proc. IEEE Int. Conf. Robot. Autom. (ICRA)},
  year      = {2025}
}

@inproceedings{yang2025opengsfusion,
  title     = {OpenGS-Fusion: Open-Vocabulary Dense Mapping with Hybrid {3D} Gaussian Splatting for Refined Object-Level Understanding},
  author    = {Yang, Dianyi and Wang, Xihan and Gao, Yu and Liu, Shiyang and Ren, Bohan and Yue, Yufeng and Yang, Yi},
  booktitle = {Proc. IEEE/RSJ Int. Conf. Intell. Robots Syst. (IROS)},
  year      = {2025}
}

@inproceedings{ha2024rgbd,
  title     = {{RGBD} {GS-ICP} {SLAM}},
  author    = {Ha, Seongbo and Yeon, Jiung and Yu, Hyeonwoo},
  booktitle = {Proc. Eur. Conf. Comput. Vis. (ECCV)},
  year      = {2024}
}

@inproceedings{mairal2009online,
  title     = {Online Dictionary Learning for Sparse Coding},
  author    = {Mairal, Julien and Bach, Francis and Ponce, Jean and Sapiro, Guillermo},
  booktitle = {Proc. Int. Conf. Mach. Learn. (ICML)},
  year      = {2009}
}

@article{kerbl20233d,
  title   = {{3D} Gaussian Splatting for Real-Time Radiance Field Rendering},
  author  = {Kerbl, Bernhard and Kopanas, Georgios and Leimk{\"u}hler, Thomas and Drettakis, George},
  journal = {ACM Trans. Graph.},
  year    = {2023}
}

@article{titkov2025leg,
  title   = {{LEG}-{SLAM}: Real-Time Language-Enhanced Gaussian Splatting for {SLAM}},
  author  = {Titkov, Roman and Zubkov, Egor and Yudin, Dmitry and Mahmoud, Jaafar and Mohrat, Malik and Sidorov, Gennady},
  journal = {arXiv preprint arXiv:2506.03073},
  year    = {2025}
}

@article{lee2025lego,
  title   = {{LEGO}-{SLAM}: Language-Embedded Gaussian Optimization {SLAM}},
  author  = {Lee, Sibaek and Ha, Seongbo and Kang, Kyeongsu and Choi, Joonyeol and Tak, Seungjun and Yu, Hyeonwoo},
  journal = {arXiv preprint arXiv:2511.16144},
  year    = {2025}
}

@article{yoo2025openmonogs,
  title   = {OpenMonoGS-{SLAM}: Monocular Gaussian Splatting {SLAM} with Open-Set Semantics},
  author  = {Yoo, Jisang and Kang, Gyeongjin and Ko, Hyun-kyu and Yu, Hyeonwoo and Park, Eunbyung},
  journal = {arXiv preprint arXiv:2512.08625},
  year    = {2025}
}

@article{deng2025omnimap,
  title   = {OmniMap: A General Mapping Framework Integrating Optics, Geometry, and Semantics},
  author  = {Deng, Yinan and Yue, Yufeng and Dou, Jianyu and Zhao, Jingyu and Wang, Jiahui and Tang, Yujie and Yang, Yi and Fu, Mengyin},
  journal = {IEEE Trans. Robot.},
  year    = {2025}
}

@inproceedings{tian2025ccl,
  title     = {{CCL}-{LGS}: Contrastive Codebook Learning for {3D} Language Gaussian Splatting},
  author    = {Tian, Lei and Li, Xiaomin and Ma, Liqian and Yin, Hao and Zheng, Zirui and Huang, Hefei and Li, Taiqing and Lu, Huchuan and Jia, Xu},
  booktitle = {Proc. IEEE/CVF Int. Conf. Comput. Vis. (ICCV)},
  year      = {2025}
}

@inproceedings{qu2024goi,
  title     = {{GOI}: Find {3D} Gaussians of Interest with an Optimizable Open-Vocabulary Semantic-Space Hyperplane},
  author    = {Qu, Yansong and Dai, Shaohui and Li, Xinyang and Lin, Jianghang and Cao, Liujuan and Zhang, Shengchuan and Ji, Rongrong},
  booktitle = {Proc. ACM Int. Conf. Multimedia (ACM MM)},
  year      = {2024}
}

@article{campos2021orb,
  title   = {{ORB}-{SLAM3}: An Accurate Open-Source Library for Visual, Visual--Inertial, and Multi-Map {SLAM}},
  author  = {Campos, Carlos and Elvira, Richard and Rodr{\'\i}guez, Juan J. G{\'o}mez and Montiel, Jos{\'e} M. M. and Tard{\'o}s, Juan D.},
  journal = {IEEE Trans. Robot.},
  year    = {2021},
  publisher = {IEEE}
}

@inproceedings{o2002streaming,
  title     = {Streaming-Data Algorithms for High-Quality Clustering},
  author    = {O'Callaghan, Liadan and Mishra, Nina and Meyerson, Adam and Guha, Sudipto and Motwani, Rajeev},
  booktitle = {Proc. Int. Conf. Data Eng. (ICDE)},
  year      = {2002},
}

@article{straub2019replica,
  title   = {The Replica Dataset: A Digital Replica of Indoor Spaces},
  author  = {Straub, Julian and Whelan, Thomas and Ma, Lingni and Chen, Yufan and Wijmans, Erik and Green, Simon and Engel, Jakob J. and Mur-Artal, Raul and Ren, Carl and Verma, Shobhit and others},
  journal = {arXiv preprint arXiv:1906.05797},
  year    = {2019}
}

@inproceedings{sturm2012benchmark,
  title     = {A Benchmark for the Evaluation of {RGB-D} {SLAM} Systems},
  author    = {Sturm, J{\"u}rgen and Engelhard, Nikolas and Endres, Felix and Burgard, Wolfram and Cremers, Daniel},
  booktitle = {Proc. IEEE/RSJ Int. Conf. Intell. Robots Syst. (IROS)},
  year      = {2012}
}

@article{tang2023mips,
  title   = {MIPS-Fusion: Multi-Implicit-Submaps for Scalable and Robust Online Neural {RGB-D} Reconstruction},
  author  = {Tang, Yijie and Zhang, Jiazhao and Yu, Zhinan and Wang, He and Xu, Kai},
  journal = {ACM Trans. Graph.},
  year    = {2023},
  publisher = {ACM}
}

@inproceedings{zhang2021invertible,
  title     = {Invertible Concept-Based Explanations for {CNN} Models with Non-Negative Concept Activation Vectors},
  author    = {Zhang, Ruihan and Madumal, Prashan and Miller, Tim and Ehinger, Krista A. and Rubinstein, Benjamin I. P.},
  booktitle = {Proc. AAAI Conf. Artif. Intell. (AAAI)},
  year={2021},
}

@article{touvron2023llama,
  title   = {{LLaMA}: Open and Efficient Foundation Language Models},
  author  = {Touvron, Hugo and Lavril, Thibaut and Izacard, Gautier and Martinet, Xavier and Lachaux, Marie-Anne and Lacroix, Timoth{\'e}e and Rozi{\`e}re, Baptiste and Goyal, Naman and Hambro, Eric and Azhar, Faisal and others},
  journal = {arXiv preprint arXiv:2302.13971},
  year    = {2023}
}

@inproceedings{conceptfusion,
  title     = {ConceptFusion: Open-Set Multimodal {3D} Mapping},
  author    = {Jatavallabhula, Krishna Murthy and Kuwajerwala, Alihusein and Gu, Qiao and Omama, Mohd and Chen, Tao and Li, Shuang and Iyer, Ganesh and Saryazdi, Soroush and Keetha, Nikhil and Tewari, Ayush and Tenenbaum, Joshua B. and de Melo, Celso Miguel and Krishna, Madhava and Paull, Liam and Shkurti, Florian and Torralba, Antonio},
  booktitle = {Proc. Robot.: Sci. Syst. (RSS)},
  year      = {2023}
}

@inproceedings{weijler2025openhype,
  title     = {OpenHype: Hyperbolic Embeddings for Hierarchical Open-Vocabulary Radiance Fields},
  author    = {Weijler, L. and Koch, S. and Poiesi, F. and Ropinski, T. and Hermosilla, P.},
  booktitle = {Adv. Neural Inf. Process. Syst. (NeurIPS)},
  year      = {2025}
}

@article{tosi2024nerfs,
  title   = {How {NeRF}s and {3D} Gaussian Splatting Are Reshaping {SLAM}: A Survey},
  author  = {Tosi, Fabio and Zhang, Youmin and Gong, Ziren and Sandstr{\"o}m, Erik and Mattoccia, Stefano and Oswald, Martin R. and Poggi, Matteo},
  journal = {arXiv preprint arXiv:2402.13255},
  year    = {2024}
}

@inproceedings{keetha2024splatam,
  title     = {SplaTAM: Splat Track \& Map {3D} Gaussians for Dense {RGB-D} {SLAM}},
  author    = {Keetha, Nikhil and Karhade, Jay and Jatavallabhula, Krishna Murthy and Yang, Gengshan and Scherer, Sebastian and Ramanan, Deva and Luiten, Jonathon},
  booktitle = {Proc. IEEE/CVF Conf. Comput. Vis. Pattern Recognit. (CVPR)},
  year      = {2024}
}

@inproceedings{matsuki2024gaussian,
  title     = {Gaussian Splatting {SLAM}},
  author    = {Matsuki, Hidenobu and Murai, Riku and Kelly, Paul H. J. and Davison, Andrew J.},
  booktitle = {Proc. IEEE/CVF Conf. Comput. Vis. Pattern Recognit. (CVPR)},
  year      = {2024}
}

@article{thirgood2026featureslam,
  title={FeatureSLAM: Feature-enriched 3D gaussian splatting SLAM in real time},
  author={Thirgood, Christopher and Mendez, Oscar and Ling, Erin and Storey, Jon and Hadfield, Simon},
  journal={arXiv preprint arXiv:2601.05738},
  year={2026}
}

@article{mobile_sam,
  title={Faster Segment Anything: Towards Lightweight SAM for Mobile Applications},
  author={Zhang, Chaoning and Han, Dongshen and Qiao, Yu and Kim, Jung Uk and Bae, Sung-Ho and Lee, Seungkyu and Hong, Choong Seon},
  journal={arXiv preprint arXiv:2306.14289},
  year={2023}
}

@ARTICLE{wilson2025latent,
  author={Wilson, Joey and Xu, Ruihan and Sun, Yile and Ewen, Parker and Zhu, Minghan and Barton, Kira and Ghaffari, Maani},
  journal={IEEE Robotics and Automation Letters}, 
  title={{LatentBKI}: Open-Dictionary Continuous Mapping in Visual-Language Latent Spaces With Quantifiable Uncertainty}, 
  year={2025}
}

@inproceedings{shi2024language,
  title={Language embedded 3d gaussians for open-vocabulary scene understanding},
  author={Shi, Jin-Chuan and Wang, Miao and Duan, Hao-Bin and Guan, Shao-Hua},
  booktitle={Proc. IEEE/CVF Conf. Comput. Vis. Pattern Recognit. (CVPR)},
  year={2024}
}

@article{zuo2025fmgs,
  title={Fmgs: Foundation model embedded 3d gaussian splatting for holistic 3d scene understanding},
  author={Zuo, Xingxing and Samangouei, Pouya and Zhou, Yunwen and Di, Yan and Li, Mingyang},
  journal={International Journal of Computer Vision},
  year={2025}
}

@inproceedings{qin2024langsplat,
  title={Langsplat: 3d language gaussian splatting},
  author={Qin, Minghan and Li, Wanhua and Zhou, Jiawei and Wang, Haoqian and Pfister, Hanspeter},
  booktitle={Proc. IEEE/CVF Conf. Comput. Vis. Pattern Recognit. (CVPR)},
  year={2024}
}

@inproceedings{deng2025openvox,
  title={Openvox: Real-time instance-level open-vocabulary probabilistic voxel representation},
  author={Deng, Yinan and Yao, Bicheng and Tang, Yihang and Zhou, Tianxing and Yang, Yi and Yue, Yufeng},
  booktitle={Proc. IEEE/RSJ Int. Conf. Intell. Robots Syst. (IROS)},
  year={2025},
}

@inproceedings{jin2025openfusion++,
  title={OpenFusion++: An Open-vocabulary Real-time Scene Understanding System},
  author={Jin, Xiaofeng and Frosi, Matteo and Matteucci, Matteo},
  booktitle={Proc. IEEE/RSJ Int. Conf. Intell. Robots Syst. (IROS)},
  year={2025},
}
\end{document}